\documentclass[11pt]{article}
\pdfoutput=1
\usepackage{acl2016}
\usepackage{times}
\usepackage{latexsym}
\usepackage{amsmath}
\usepackage{multirow}
\usepackage{url}
\usepackage{graphicx}
\usepackage[linesnumbered,ruled,vlined,commentsnumbered]{algorithm2e}
\SetAlFnt{\small}
\usepackage{amssymb}
\usepackage{amsfonts}
\usepackage{todonotes}
\usepackage[hidelinks]{hyperref}
\usepackage{fancyvrb}
\usepackage{tikz}
\usetikzlibrary{bayesnet}
\usepackage {xcolor}
\usetikzlibrary {positioning}
\usepackage[inline]{enumitem}

\usepackage{xcolor}
\newcommand{\CHANGEB}[1]{#1}   

\aclfinalcopy 
\def\aclpaperid{92} 

\title{Neighborhood Mixture Model for Knowledge Base Completion\thanks{\ \ A revised version of our CoNLL 2016 paper to update latest related work.}}

\author{Dat Quoc Nguyen${}^{1}$, Kairit Sirts${}^{1}$, Lizhen Qu${}^{2}$ \and Mark Johnson${}^{1}$ \\
\\
${}^{1}$ Department of Computing, Macquarie University, Sydney, Australia \\
{\tt{\small{dat.nguyen@students.mq.edu.au, \{kairit.sirts, mark.johnson\}@mq.edu.au}}} \\
${}^{2}$ Data61 \& Australian National University \\
{\tt{\small{lizhen.qu@data61.csiro.au}}}
}

\begin{document}

\maketitle

\begin{abstract}
Knowledge bases are useful resources for many  natural language processing tasks, however, they are far from complete. In this paper, we define a novel entity representation as a mixture of its neighborhood in the knowledge base and apply this technique on TransE---a well-known embedding model for knowledge base completion. Experimental results show that the neighborhood information significantly helps to improve the results of the TransE model, leading to better performance than obtained by other state-of-the-art embedding models on three benchmark datasets for  triple classification, entity prediction and relation prediction tasks. 

\end{abstract}

\section{Introduction} \label{sec:intro}
Knowledge bases (KBs), such as WordNet \cite{Miller:1995:WLD:219717.219748}, YAGO
\cite{Suchanek:2007}, Freebase \cite{Bollacker:2008} and DBpedia
\cite{LehmannIJJKMHMK15}, represent relationships between entities as
triples $(\mathrm{head\ entity, relation, tail\ entity})$.  Even very
large knowledge bases are still far from complete
\cite{NIPS2013_5028,West:2014:KBC:2566486.2568032}.
\textit{Knowledge base completion} or \textit{link prediction}  systems
\cite{NickelMTG15} predict which triples not in a knowledge base are
likely to be true  \cite{NIPS2003_2465,bordes-2011}. 

\textit{Embedding models} for KB completion associate entities and/or
relations with dense feature vectors or matrices.  Such models obtain
state-of-the-art performance
\cite{NIPS2013_5071,AAAI148531,guu-miller-liang:2015:EMNLP,ijcai/0005HZ16,NguyenNAACL2016,DasNBM17}
and generalize to large KBs \cite{Denis2015}.  


Most embedding models for KB completion learn only from triples and by doing so, ignore lots of information implicitly provided by the structure of the knowledge graph.
Recently, several authors have addressed this issue by incorporating relation path information into model learning \cite{lin-EtAl:2015:EMNLP1,guu-miller-liang:2015:EMNLP,toutanova-EtAl:2016:P16-1} and have shown that the relation paths between entities in KBs provide useful information and improve knowledge base completion.
For instance, a three-relation path
\begin{align*}
&(\mathrm{head}, \mathrm{born\_in\_hospital}/\mathrm{r_1}, \mathrm{e_1})  \\
\Rightarrow &(\mathrm{e_1}, \mathrm{hospital\_located\_in\_city}/\mathrm{r_2}, \mathrm{e_2}) \\
\Rightarrow &(\mathrm{e_2}, \mathrm{city\_in\_country}/\mathrm{r_3}, \mathrm{tail})
\end{align*}
is likely to indicate that the fact $(\mathrm{head}, \mathrm{nationality}, \mathrm{tail})$ could be true, so the relation path here $\mathrm{p} = \{\mathrm{r_1}, \mathrm{r_2}, \mathrm{r_3}\}$ is useful for predicting the relationship ``$\mathrm{nationality}$'' between the $\mathrm{head}$ and $\mathrm{tail}$ entities.

\begin{figure}[ht]
\centering
\begin{tikzpicture}[-latex ,auto ,node distance =2cm and 2cm ,on grid ,semithick ,state/.style ={ ellipse,draw,text=black , minimum width =1cm}]
\node[state] (Ben) {Ben Affleck};
\node[state] (Male) [below left =of Ben] at (-1,0.25) {male};
\path (Ben) edge node[left] {gender} (Male);
\node[state] (Actor) [below right =of Ben]  at (0.5,0.5) {actor};
\draw[dotted] (Ben) edge node[right] {occupation?} (Actor);
\node[state] (FP) [below right =of Ben]  at (-0.5,-3) {film maker};
\draw[dotted] (Ben) edge node[right] {occupation?} (FP);
\node[state] (Oscar) [below right =of Ben] at (-2.5,-4) {Oscar award};
\path (Ben) edge node[left=-3pt] {won} (Oscar);
\node[state] (LA) [below left =of Ben] at (-0.6,-3) {Los Angeles};
\path (Ben) edge node[left] {live\ in} (LA);
\node[state] (Lect) [above right =of Ben]  {lecturer};
\draw[dotted] (Ben) edge node[right] {occupation?} (Lect);
\node[state] (ViA) [above left =of Ben] at (0.5,0) {Violet Anne};
\path (ViA) edge node[left] {child\ of} (Ben);
\end{tikzpicture}
\caption{An example fragment of a KB.}
\label{fig:neigh}
\end{figure}
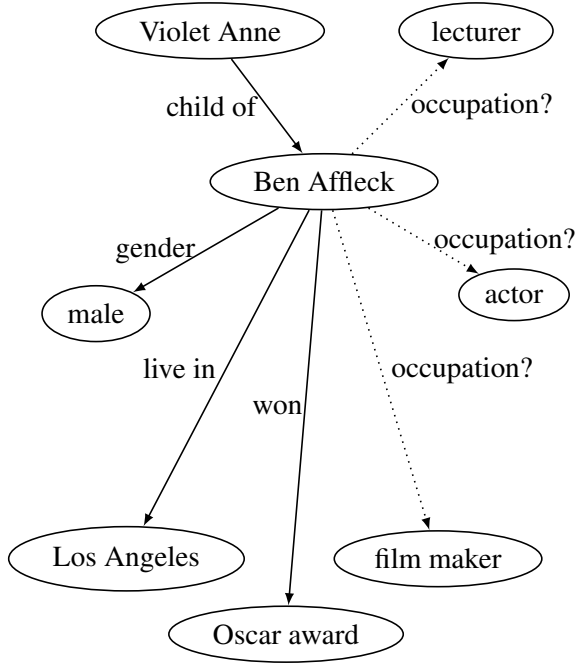

Besides the relation paths, there could be other useful information implicitly presented in the knowledge base that could be exploited for better KB completion. For instance, the whole neighborhood of entities could provide lots of useful information for predicting the relationship between two entities. Consider for example a KB fragment given in Figure~\ref{fig:neigh}. If we know that Ben Affleck has won an Oscar award and Ben Affleck lives in Los\ Angeles, then this can help us to predict that Ben Affleck is an actor or a film maker, rather than a lecturer or a doctor. If we additionally know that Ben Affleck's gender is male then there is a higher chance for him to be a film maker. This intuition can be formalized by representing an entity vector as a relation-specific mixture of its neighborhood as follows:

\setlength{\abovedisplayskip}{-5pt}
\setlength{\belowdisplayskip}{-5pt}
\begin{align*}
\mathrm{Ben\_Affleck} &= \omega_{r,1} (\mathrm{Violet\_Anne}, \mathrm{child\_of}) \\
		     &+ \omega_{r,2} (\mathrm{male}, \mathrm{gender}^{-1}) \\
		     &+ \omega_{r,3} (\mathrm{Los\_Angeles}, \mathrm{lives\_in}^{-1}) \\
		     &+ \omega_{r,4} (\mathrm{Oscar\_award}, \mathrm{won}^{-1}), \\
\end{align*}

\noindent where $\omega_{r,i}$ are the mixing weights that indicate how important each neighboring relation is for predicting the relation $r$. For example, for predicting the $\mathrm{occupation}$ relationship, the knowledge about the  $\mathrm{child\_of}$ relationship might not be that informative and thus the corresponding mixing coefficient can be close to zero, whereas it could be relevant for predicting some other relationship, such as $\mathrm{parent}$ or $\mathrm{spouse}$, in which case the relation-specific mixing coefficient for the $\mathrm{child\_of}$ relationship could be high.

The primary contribution of this paper is introducing and formalizing the neighborhood mixture model. We demonstrate its usefulness by applying it to the well-known TransE model \cite{NIPS2013_5071}. However, it could be applied to other embedding models as well, such as {Bilinear} models \cite{Bordes2014SME,yang-etal-2015}  and STransE \cite{NguyenNAACL2016}. While relation path models exploit extra information using longer paths existing in the KB, the neighborhood mixture model effectively incorporates information about many paths simultaneously. Our extensive experiments on three benchmark datasets show that it achieves superior performance over competitive baselines in three KB completion tasks: triple classification, entity prediction and relation prediction. 

\section{Neighborhood mixture modeling}
%
In this section, we start by explaining how to formally construct the neighbor-based entity representations in section~\ref{sec:nber}, and then describe the \textbf{N}eighborhood \textbf{M}ixture \textbf{M}odel applied to the TransE model \cite{NIPS2013_5071} in section~\ref{sec:transe_nmm}. Section~\ref{sec:par_opt} explains how we train our model.

\subsection{Neighbor-based entity representation}
\label{sec:nber}
Let $\mathcal{E}$ denote the set of entities and $\mathcal{R}$ the set of relation types. Denote by $\mathcal{R}^{-1}$ the set of inverse relations $r^{-1}$. Denote by $\mathcal{G}$ the knowledge graph consisting of a set of correct tiples $(h, r, t)$, such that $h, t \in \mathcal{E}$ and $r \in \mathcal{R}$. Let $\mathcal{K}$ denote the symmetric closure of $\mathcal{G}$, i.e. if a triple $(h, r, t) \in \mathcal{G}$, then both $(h, r, t)$ and $(t, r^{-1}, h) \in \mathcal{K}$.


%
%
Define:

\vspace{-10pt}
\begin{equation*} 
 \mathcal{N}_{e,r} = \{e' | (e', r, e) \in \mathcal{K}\}
 \end{equation*}
 
 \noindent as a set of neighboring entities connected to entity $e$ with relation $r$. Then  

\setlength{\belowdisplayskip}{7pt}
\begin{equation*} 
 \mathcal{N}_e = \{(e',r) | r  \in \mathcal{R} \cup \mathcal{R}^{-1} ,  e' \in \mathcal{N}_{e,r}\}
\end{equation*}

\noindent is the set of all entity and relation pairs that are neighbors for entity $e$.

Each entity $e$ is associated with a k-dimensional vector $\boldsymbol{v}_{e}\in\mathbb{R}^{k}$ and relation-dependent vectors  $\boldsymbol{u}_{e, r}\in\mathbb{R}^{k}, r \in \mathcal{R} \cup \mathcal{R}^{-1}$. 
Now we can define the neighborhood-based entity representation $\boldsymbol{\vartheta}_{e,r}$ for an entity $e \in
\mathcal{E}$ for predicting the relation $r  \in \mathcal{R}$ as follows:

\begin{equation}
\boldsymbol{\vartheta}_{e,r} = a_e\boldsymbol{v}_e + \sum_{(e',r') \in \mathcal{N}_e} b_{r,r'}\boldsymbol{u}_{e',r'},
\label{eql:V_eh}
\end{equation}

\noindent $a_e$ and $b_{r, r'}$ are the mixture weights that are constrained to sum to 1 for each neighborhood:

\begin{align}
a_{e} & \propto  \delta + \exp\alpha_e  \\
b_{r,r'} & \propto  \exp \beta_{r,r'}  
\end{align}

\noindent where $ \delta \geqslant 0$ is a hyper-parameter that controls the contribution of the entity vector $\boldsymbol{v}_{e}$ to the neighbor-based mixture, $\alpha_e$ and $\beta_{r, r'}$ are the learnable exponential mixture parameters.

In real-world factual KBs, e.g. Freebase \cite{Bollacker:2008}, some entities, such as ``$\mathrm{male}$'', can have thousands or millions neighboring entities sharing the same relation ``$\mathrm{gender}$.'' 
For such entities, computing the neighbor-based vectors can be computationally very expensive.
To overcome this problem, we introduce in our implementation a  filtering threshold $\tau$ and consider in the neighbor-based entity representation construction only those relation-specific neighboring entity sets  for which $|\mathcal{N}_{e,r}| \leq \tau$.


\subsection{TransE-NMM: applying neighborhood mixtures to TransE}
\label{sec:transe_nmm}
Embedding models define for each triple $(h, r, t) \in \mathcal{G}$, a \textit{score function} $f(h, r, t)$ that measures its implausibility.
The goal is to choose $f$ such that the score $f(h, r, t)$ of a plausible triple $(h, r, t)$ is
smaller than the score $f(h', r', t')$ of an implausible triple $(h', r', t')$.

TransE \cite{NIPS2013_5071} is a simple embedding model for knowledge base completion, which, despite of its simplicity, obtains very competitive results \cite{Garcia-DuranBUG15,Nickel:2016:HEK:3016100.3016172}. In TransE, both entities $e$ and relations $r$ are represented with k-dimensional vectors $\boldsymbol{v}_{e} \in \mathbb{R}^{k}$ and $\boldsymbol{v}_{r} \in \mathbb{R}^{k}$, respectively. These vectors are chosen such that for each triple $(h, r, t) \in \mathcal{G}$:

\begin{equation}
\boldsymbol{v}_{h} + \boldsymbol{v}_{r} \approx \boldsymbol{v}_{t}
\end{equation}

\noindent The score function of the TransE model is the norm of this translation:

\begin{equation}
f(h, r, t)_{\mathrm{TransE}} =  \| \boldsymbol{v}_{h} + \boldsymbol{v}_{r} - \boldsymbol{v}_{t} \|_{\ell_{1/2}} 
\end{equation}

\noindent We define the score function of our new model TransE-NMM in terms of the neighbor-based entity vectors as follows:

\setlength{\abovedisplayskip}{-5pt}
\setlength{\belowdisplayskip}{10pt}
\begin{equation}
f(h, r, t) =  \| \boldsymbol{\vartheta}_{h,r} + \boldsymbol{v}_{r} - \boldsymbol{\vartheta}_{t,r^{-1}} \|_{\ell_{1/2}} ,
\end{equation}

\noindent using either  the $\ell_1$ or the $\ell_2$-norm, and  $\boldsymbol{\vartheta}_{h,r}$ and  $\boldsymbol{\vartheta}_{t,r^{-1}}$ are defined following the Equation \ref{eql:V_eh}. The relation-specific entity vectors $\boldsymbol{u}_{e, r}$ used to construct the neighbor-based entity vectors $\boldsymbol{\vartheta}_{e,r}$ are defined based on the TransE translation operator:

\setlength{\abovedisplayskip}{-5pt}
\setlength{\belowdisplayskip}{10pt}
\begin{equation}
\boldsymbol{u}_{e,r} = \boldsymbol{v}_e +  \boldsymbol{v}_{r}\label{equal:uer}
\end{equation}

\noindent in which $\boldsymbol{v}_{r^{-1}} = -\boldsymbol{v}_{r}$. For each correct triple $(h, r, t)$, the sets of neighboring entities $\mathcal{N}_{h,r}$ and  $\mathcal{N}_{t,r^{-1}}$ exclude the entities $t$ and $h$, respectively. 

If we set the filtering threshold $\tau = 0$ then  $\boldsymbol{\vartheta}_{h,r} = \boldsymbol{v}_{h}$ and $\boldsymbol{\vartheta}_{t,r^{-1}} = \boldsymbol{v}_{t}$ for all triples.  In this case, TransE-NMM reduces to the plain TransE model.  In all our experiments presented in  section \ref{sec:exp}, the baseline TransE results are obtained with the TransE-NMM  with $\tau = 0$. 

\subsection{Parameter optimization}
\label{sec:par_opt}

The TransE-NMM model parameters include the   vectors $\boldsymbol{v}_{e}, \boldsymbol{v}_{r}$ for entities and relation types, the entity-specific weights  
$\boldsymbol{\alpha} = \{\alpha_e | e \in \mathcal{E}\}$ and relation-specific weights 
$\boldsymbol{\beta} = \{\beta_{r,r'} | r,r' \in \mathcal{R} \cup \mathcal{R}^{-1} \}$. To learn these parameters, we minimize the $L_2$-regularized margin-based objective function:

\begin{align}
\mathcal{L} & =  \sum_{\substack{(h,r,t) \in \mathcal{G} \\ (h',r,t') \in \mathcal{G}'_{(h, r, t)}}} [\gamma + f(h, r, t) - f(h', r, t')]_+  \nonumber \\
 & +  \dfrac{\lambda}{2} \Big(\|\boldsymbol{\alpha}\|_2^2 + \|\boldsymbol{\beta}\|_2^2 \Big),
 \label{equal:objfunc}
  \end{align}

\noindent where $[x]_+ = \max(0, x)$,  $\gamma$ is the margin hyper-parameter, $\lambda$ is the $L_2$ regularization parameter and

\begin{align*}
\mathcal{G}'_{(h, r, t)} &= \lbrace (h', r, t) \mid h' \in
\mathcal{E}, (h', r, t) \notin \mathcal{G} \rbrace \\
& \cup \lbrace (h, r,
t') \mid t' \in \mathcal{E}, (h, r, t') \notin \mathcal{G} \rbrace
\end{align*}

\noindent is the set of incorrect triples generated by corrupting the
correct triple $(h, r, t)\in\mathcal{G}$. We applied the ``\textit{Bernoulli}'' trick to choose whether to generate  
the head or tail entities when sampling  incorrect triples \cite{AAAI148531,AAAI159571,He:2015,ji-EtAl:2015:ACL-IJCNLP,JiLH016}.


We use Stochastic
Gradient Descent (SGD) with RMSProp adaptive learning rate to minimize $\mathcal{L}$, and impose the following hard constraints during training: $\|\boldsymbol{v}_{e}\|_2 \leqslant 1$ and $\|\boldsymbol{v}_{r}\|_2 \leqslant 1$.
 We employ  alternating optimization  to minimize $\mathcal{L}$. 
 We first initialize the entity and relation-specific mixing parameters $\boldsymbol{\alpha}$ and $\boldsymbol{\beta}$ to zero and only learn the randomly initialized entity and relation vectors $\boldsymbol{v}_{e}$ and $\boldsymbol{v}_{r}$.
 Then we fix the learned vectors and only optimize the mixing parameters. 
 In the final step, we fix again the mixing parameters and fine-tune the vectors.
 In all experiments presented in section \ref{sec:exp}, we train for 200 epochs during each three optimization step.

\begin{table*}[!ht]
\centering
\resizebox{16cm}{!}{
\def\arraystretch{1.125}
\begin{tabular}{l|l|l}
\hline
\bf Model & Score function $f(h, r, t)$ & \bf Opt.  \\
\hline
\hline
STransE & $\| \textbf{W}_{r,1}\boldsymbol{v}_{h} + \boldsymbol{v}_{r} - \textbf{W}_{r,2}\boldsymbol{v}_{t}\|_{\ell_{1/2}}$ ;  $\textbf{W}_{r,1}$, $\textbf{W}_{r,2}$ $\in$ $\mathbb{R}^{k \times k}$; $\boldsymbol{v}_{r} \in \mathbb{R}^{k}$ & SGD  \\
\hline
SE &  $\| \textbf{W}_{r,1}\boldsymbol{v}_{h} - \textbf{W}_{r,2}\boldsymbol{v}_{t}\|_{\ell_{1/2}}$ ;  $\textbf{W}_{r,1}$, $\textbf{W}_{r,2}$ $\in$ $\mathbb{R}^{k \times k}$ & SGD\\
\hline
Unstructured &  $ \| \boldsymbol{v}_{h} - \boldsymbol{v}_{t} \|_{\ell_{1/2}}$ & SGD \\
\hline
TransE &  $ \|  \boldsymbol{v}_{h} + \boldsymbol{v}_{r} - \boldsymbol{v}_{t} \|_{\ell_{1/2}}$ ;   $\boldsymbol{v}_{r}  \in  \mathbb{R}^{k}$ & SGD \\
\hline
\multirow{2}{*}{TransH} & $\| (\textbf{I} - \boldsymbol{r}_{p}\boldsymbol{r}_{p}^{\top})\boldsymbol{v}_{h} + \boldsymbol{v}_{r} - (\textbf{I} - \boldsymbol{r}_{p}\boldsymbol{r}_{p}^{\top})\boldsymbol{v}_{t} \|_{\ell_{1/2}}$  & \multirow{2}{*}{SGD}\\
&  $\boldsymbol{r}_{p}$, $\boldsymbol{v}_{r} \in$ $\mathbb{R}^{k}$ ; $\textbf{I}$: Identity matrix size $k \times k$ & \\
\hline
\multirow{2}{*}{TransD} & $\| (\textbf{I} + \boldsymbol{r}_{p}\boldsymbol{h}_{p}^{\top})\boldsymbol{v}_{h} + \boldsymbol{v}_{r} - (\textbf{I} + \boldsymbol{r}_{p}\boldsymbol{t}_{p}^{\top})\boldsymbol{v}_{t} \|_{\ell_{1/2}}$ & \multirow{2}{*}{AdaDelta} \\
&    $\boldsymbol{r}_{p}$, $\boldsymbol{v}_{r}$ $\in$ $\mathbb{R}^{n}$ ; $\boldsymbol{h}_{p}, \boldsymbol{t}_{p}$ $\in$ $\mathbb{R}^{k}$ ; $\textbf{I}$: Identity matrix size $n \times k$  & \\
\hline
TransR & $\| \textbf{W}_{r}\boldsymbol{v}_{h} + \boldsymbol{v}_{r} - \textbf{W}_{r}\boldsymbol{v}_{t}\|_{\ell_{1/2}}$ ;  $\textbf{W}_{r}$ $\in$ $\mathbb{R}^{n \times k}$  ;    $\boldsymbol{v}_{r}$ $\in$ $\mathbb{R}^{n}$ & SGD \\
\hline
TranSparse & $\| \textbf{W}_{r}^h(\theta_r^h)\boldsymbol{v}_{h} + \boldsymbol{v}_{r} - \textbf{W}_{r}^t(\theta_r^t)\boldsymbol{v}_{t}\|_{\ell_{1/2}}$ ;  $\textbf{W}_{r}^h$, $\textbf{W}_{r}^t$ $\in$ $\mathbb{R}^{n \times k}$; $\theta_r^h$, $\theta_r^t \in \mathbb{R}$ ; $\boldsymbol{v}_{r}$ $\in$ $\mathbb{R}^{n}$ & SGD  \\
\hline
\multirow{2}{*}{SME} & $(\textbf{W}_{1,1}\boldsymbol{v}_{h} + \textbf{W}_{1,2}\boldsymbol{v}_{r} + \textbf{b}_1)^{\top}(\textbf{W}_{2,1}\boldsymbol{v}_{t} + \textbf{W}_{2,2}\boldsymbol{v}_{r} + \textbf{b}_2)$   & \multirow{2}{*}{SGD}  \\
 &  $\textbf{b}_1\text{, } \textbf{b}_2 \in \mathbb{R}^n$; $\textbf{W}_{1,1}$, $\textbf{W}_{1,2}, \textbf{W}_{2,1}$, $\textbf{W}_{2,2} \in \mathbb{R}^{n \times k}$ & \\
\hline
DISTMULT &  $ \boldsymbol{v}_{h}^{\top}\textbf{W}_{r}\boldsymbol{v}_{t}$ ;   $\textbf{W}_{r}$ is a diagonal matrix $\in$ $\mathbb{R}^{k \times k}$ & AdaGrad\\
\hline
\multirow{2}{*}{NTN} & $ \boldsymbol{v}_r^{\top} \mathit{tanh}( \boldsymbol{v}_{h}^{\top}\textbf{M}_{r}\boldsymbol{v}_{t}  + \textbf{W}_{r,1}\boldsymbol{v}_{h} + \textbf{W}_{r,2}\boldsymbol{v}_{t} + \textbf{b}_r)$   & \multirow{2}{*}{L-BFGS}  \\
 &  $\boldsymbol{v}_r\text{, } \textbf{b}_r \in \mathbb{R}^n$; $\textbf{M}_{r} \in \mathbb{R}^{k \times k \times n}$;  $\textbf{W}_{r,1}$, $\textbf{W}_{r,2} \in \mathbb{R}^{n \times k}$ & \\
\hline
Bilinear-\textsc{comp} &  $ \boldsymbol{v}_{h}^{\top}\textbf{W}_{r_1}\textbf{W}_{r_2}...\textbf{W}_{r_m}\boldsymbol{v}_{t}$\ ; $\textbf{W}_{r_1}, \textbf{W}_{r_2},..., \textbf{W}_{r_m} \in \mathbb{R}^{k \times k}$ & AdaGrad\\
\hline
TransE-\textsc{comp} &  $ \|  \boldsymbol{v}_{h} + \boldsymbol{v}_{r_1} + \boldsymbol{v}_{r_2} + ... + \boldsymbol{v}_{r_m} - \boldsymbol{v}_{t} \|_{\ell_{1/2}}$ ;     $\boldsymbol{v}_{r_1}, \boldsymbol{v}_{r_2} ,..., \boldsymbol{v}_{r_m} \in \mathbb{R}^{k}$ & AdaGrad\\
\hline
\end{tabular}
}
\caption{The score functions $f(h,r, t)$ and the optimization methods
  (Opt.) of several prominent embedding models for KB completion.  
  In all of these models, the entities $h$ and $t$ are represented
  by  vectors $\boldsymbol{v}_{h}$ and $\boldsymbol{v}_{t} \in \mathbb{R}^{k}$ respectively.
}
\label{tab:emmethods}
\end{table*}

\section{Related work}
\label{sec:related}
Table \ref{tab:emmethods} summarizes related embedding models for link
prediction and KB completion.  The models differ in their score
function $f(h, r, t)$ and the algorithm used to optimize their
margin-based objective function, e.g., SGD, AdaGrad
\cite{Duchi:2011:ASM:1953048.2021068}, AdaDelta
\cite{DBLP:journals/corr/abs-1212-5701} or L-BFGS \cite{Liu1989}.

The \textit{Unstructured} model \cite{Bordes2014SME} assumes that the head and tail entity vectors are similar. As the Unstructured model does not take the relationship into account, it cannot distinguish different relation types. The \textit{Structured Embedding} (SE) model \cite{bordes-2011} extends the Unstructured model by assuming that the head and tail entities are similar only in a relation-dependent subspace, where each relation is represented by two different matrices. Futhermore, the SME model \cite{Bordes2014SME} uses four different matrices to project entity and relation vectors into a subspace.  
The TransH model \cite{AAAI148531} associates each relation with a
relation-specific hyperplane and uses a projection vector to project
entity vectors onto that hyperplane. TransD
\cite{ji-EtAl:2015:ACL-IJCNLP} and TransR/CTransR \cite{AAAI159571}
extend the TransH model by using two projection vectors and a matrix to project entity vectors into a relation-specific space, respectively. TEKE\_H  \cite{DBLP:conf/ijcai/WangL16}  extends TransH to  incorporate rich context information in an external text corpus.  
lppTransD   \cite{yoon-EtAl:2016:N16-1}   extends the TransD model to additionally use two projection vectors for representing each relation. STransE \cite{NguyenNAACL2016} and TranSparse \cite{JiLH016} can be also viewed as extensions of the TransR model, where head and tail entities are associated with their own projection matrices. 

The DISTMULT model \cite{yang-etal-2015} is based on the \textit{Bilinear} model \cite{ICML2011Nickel_438,Bordes2014SME,NIPS2012_4744} where each relation is represented by a diagonal rather than a full matrix. The neural tensor network (NTN) model \cite{NIPS2013_5028} uses a bilinear tensor operator to represent each relation while the ProjE model \cite{ShiW16a} could be viewed as a simplified version of NTN with diagonal matrices. 
Quadratic forms are also used to model entities and relations in  KG2E \cite{He:2015}, ComplEx \cite{TrouillonWRGB16}, TATEC \cite{Garcia-DuranBUG15} and RSTE \cite{Tay:2017:RST:3018661.3018695}. In addition,  HolE \cite{Nickel:2016:HEK:3016100.3016172} uses  circular correlation---a compositional operator---which could be interpreted as a compression of the tensor product. 

Recently, several authors have shown that relation paths between entities in KBs provide richer information and improve the relationship prediction \cite{neelakantan-roth-mccallum:2015:ACL-IJCNLP,lin-EtAl:2015:EMNLP1,garciaduran-bordes-usunier:2015:EMNLP,guu-miller-liang:2015:EMNLP,wang-EtAl:2016:P16-13,feng-EtAl:2016:COLING1,Liu:2016:HRW:2911451.2911509,NIPS2016_6098,wei-zhao-liu:2016:EMNLP2016,toutanova-EtAl:2016:P16-1}. 
In fact, our TransE-NMM model can be also viewed as a three-relation path model as it takes into account the neighborhood entity and relation information of both head and tail entities in each triple.

\newcite{luo-EtAl:2015:EMNLP3} constructed relation paths between entities and viewing entities and relations in the path as pseudo-words applied  Word2Vec algorithms \cite{Mikolov13b} to produce pre-trained vectors for these pseudo-words. 
\newcite{luo-EtAl:2015:EMNLP3} showed that using these pre-trained vectors for initialization helps to improve the performance of  the TransE, SME and SE models.   \textsc{r}TransE \cite{garciaduran-bordes-usunier:2015:EMNLP}, PTransE \cite{lin-EtAl:2015:EMNLP1} and TransE-\textsc{comp} \cite{guu-miller-liang:2015:EMNLP} are extensions of the TransE model. These models similarly represent a relation path by a vector which is the sum of the vectors of all relations in the path,  whereas in the Bilinear-\textsc{comp} model \cite{guu-miller-liang:2015:EMNLP}, each relation is a matrix and so it represents the relation path by matrix multiplication. Our neighborhood mixture model can be adapted to both relation path models Bilinear-\textsc{comp} and TransE-\textsc{comp}, by replacing head and tail entity vectors by the neighbor-based vector representations, thus combining advantages of both path and neighborhood information. 
\newcite{NickelMTG15} reviews other
approaches for learning from KBs and multi-relational data.

\section{Experiments}\label{sec:exp}
To investigate the usefulness of the neighbor mixtures, we compare the  performance of the TransE-NMM against the results of the baseline TransE  and other state-of-the-art embedding models on the triple classification, entity prediction and relation prediction tasks.


\subsection{Datasets}

\begin{table}[ht]
\centering
\begin{tabular}{l|lll}
\hline
\bf Dataset: &  WN11 &  FB13 &  NELL186 \\
\hline
\#R & 11 & 13  & 186 \\
\#E &  38,696 & 75,043 & 14,463\\
\#Train & 112,581  & 316,232 & 31,134 \\
\#Valid & 2,609 & 5,908 & 5,000 \\
\#Test & 10,544 & 23,733 & 5,000 \\
\hline
\end{tabular}
\caption{Statistics of the experimental datasets used in this study
  (and \textit{previous works}). \#E is the number of entities, \#R is the
  number of relation types, and \#Train, \#Valid and \#Test are the
  numbers of correct triples in the training, validation and test sets,
  respectively. Each validation and test set also contains the same number of incorrect triples as the number of correct triples.}
\label{tab:datasets}
\end{table}

We conduct experiments using three publicly available datasets WN11, FB13 and NELL186.  For all of them, the validation and test sets containing both correct and incorrect triples have already been constructed. Statistical information about these  datasets is given in Table~\ref{tab:datasets}.

The two benchmark datasets\footnote{{\scriptsize \url{http://cs.stanford.edu/people/danqi/data/nips13-dataset.tar.bz2}}}, WN11 and FB13, were produced by \newcite{NIPS2013_5028} for triple classification. 
 WN11   is derived from the large lexical KB WordNet \cite{Miller:1995:WLD:219717.219748} involving 11 relation types.
 FB13 is derived from the large real-world fact KB FreeBase \cite{Bollacker:2008} covering 13 relation types.
The NELL186 dataset\footnote{\url{http://aclweb.org/anthology/attachments/P/P15/P15-1009.Datasets.zip}} was introduced by \newcite{guo-EtAl:2015:ACL-IJCNLP1} for both triple classification and entity prediction tasks, containing 186 most frequent relations in the KB of the CMU Never Ending Language Learning project \cite{Carlson:2010:TAN:2898607.2898816}. 


\subsection{Evaluation tasks}
We evaluate our model on three commonly used benchmark tasks: triple classification, entity prediction and relation prediction. This subsection describes those tasks in detail.

\paragraph{Triple classification:}
The triple classification task  was first introduced by \newcite{NIPS2013_5028}, and since then it has been used to evaluate various embedding models. The aim of the task is to predict whether a triple $(h, r, t)$ is correct or not. 

For classification, we set a relation-specific threshold $\theta_r$ for each relation type $r$. 
If the implausibility score of an unseen test triple  $(h, r, t)$ is smaller than $\theta_r$ then the triple will be classified as correct, otherwise incorrect.
Following \newcite{NIPS2013_5028}, the relation-specific thresholds are determined by maximizing the micro-averaged accuracy, which is a per-triple average, on the validation set. We also report the macro-averaged accuracy, which is a per-relation average.

\paragraph{Entity prediction:}
The entity prediction task
\cite{NIPS2013_5071} \CHANGEB{predicts} the head or the tail entity given the relation type and
the other \CHANGEB{entity}, i.e. predicting $h$ given $(?, r, t)$ or predicting
$t$ given $(h, r, ?)$ where $?$ denotes the missing element.
The results are evaluated using a ranking  induced by the function $f(h, r, t)$ on test triples. Note that the incorrect triples in the validation and test sets are not used for evaluating the entity prediction task nor the relation prediction task. 

Each correct test triple $(h, r, t)$ is corrupted by replacing either its head or tail entity by each of the possible 
entities in turn, and then we rank these candidates in ascending order of
their implausibility score. 
This is called as the ``Raw'' setting protocol. 
For the ``Filtered'' setting protocol described in \newcite{NIPS2013_5071}, 
 we also filter out before ranking any corrupted triples
that appear in the KB. Ranking a corrupted triple appearing in the KB  (i.e. a correct triple)  higher than the original test triple is also correct, but is penalized by the ``Raw'' score, thus the ``Filtered'' setting provides a clearer view on the ranking performance.

In addition to the mean rank and  the Hits@10 (i.e., the proportion of
test triples for which the target entity was ranked in
the top 10 predictions), which were originally used in the entity prediction task \cite{NIPS2013_5071}, we also report the mean reciprocal rank (\textbf{MRR}), which is commonly used in information retrieval. In both ``Raw'' and ``Filtered'' settings, mean rank is always greater or equal to 1  and 
lower mean
rank indicates better entity prediction performance. The MRR and Hits@10 scores always range from 0.0 to 1.0, and higher score reflects better prediction result.

\paragraph{Relation prediction:}

The relation prediction task \cite{lin-EtAl:2015:EMNLP1}
predicts the relation type given the head and tail entities, i.e. predicting $r$ given $(h, ?, t)$ where $?$ denotes the missing element.
 We corrupt each correct test triple $(h, r, t)$ by replacing its relation $r$ by each possible 
relation type in turn, and then rank these candidates in ascending order of
their implausibility score.
Just as in the entity prediction task, we use two setting protocols, ``Raw'' and ``Filtered'', and evaluate on mean rank, MRR and Hits@10.


\subsection{Hyper-parameter tuning}\label{ssec:hyper}

For all evaluation tasks, results for TransE  are obtained with TransE-NMM  with the filtering threshold $\tau = 0$, while we set $\tau = 10$ for TransE-NMM.

For triple classification, we first performed a grid search to choose the optimal hyper-parameters for TransE by monitoring the micro-averaged  triple classification  accuracy after each training epoch on the validation set. For all datasets, we chose either the $\ell_1$ or $\ell_2$ norm in the score function $f$ and the initial RMSProp learning rate $\eta \in\lbrace 0.001, 0.01 \rbrace$. Following the previous work \cite{AAAI148531,AAAI159571,ji-EtAl:2015:ACL-IJCNLP,He:2015,JiLH016},  we selected the
margin hyper-parameter $\gamma\in\lbrace 1, 2, 4 \rbrace$ and the number of vector
dimensions $k\in\lbrace 20, 50, 100 \rbrace$ on WN11 and FB13. On NELL186, we set $\gamma = 1$ and $k = 50$  \cite{guo-EtAl:2015:ACL-IJCNLP1,luo-EtAl:2015:EMNLP3}. The highest accuracy on the
\CHANGEB{validation} set was obtained when using $\eta = 0.01$ for all three datasets, and when using  $\ell_2$ norm for NELL186,  $\gamma = 4$, $k = 20$ and  $\ell_1$ norm for WN11, and $\gamma = 1$, $k = 100$ and   $\ell_2$ norm for FB13.

We set the hyper-parameters $\eta$, $\gamma$, $k$, and the $\ell_1$ or the $\ell_2$-norm in our  TransE-NMM model to the same optimal hyper-parameters searched for TransE. We then  used a grid search  to select the hyper-parameter $\delta \in \{0, 1, 5, 10\}$ and $L_2$ regularizer $\lambda \in \{0.005, 0.01, 0.05\}$ for  TransE-NMM.  By monitoring the micro-averaged accuracy after each training epoch, we obtained the highest accuracy on validation set when using $\delta = 1$ and $\lambda = 0.05$ for both WN11 and FB13, and $\delta = 0$ and $\lambda = 0.01$ for NELL186. 

For both entity prediction and relation prediction tasks, we set the hyper-parameters $\eta$, $\gamma$, $k$, and the $\ell_1$ or the $\ell_2$-norm for both TransE and  TransE-NMM to be the same as the optimal parameters found for the triple classification task. 
We then monitored on TransE the filtered MRR on validation set after each training epoch. We chose the model with highest validation MRR, which was then used to evaluate the test set.
For TransE-NMM, we searched the hyper-parameter $\delta \in \{0, 1, 5, 10\}$ and $L_2$ regularizer $\lambda \in \{0.005, 0.01, 0.05\}$. By monitoring the filtered MRR after each training epoch, we selected the best model with the highest filtered MRR   on the validation set. Specifically, for the entity prediction task, we selected $\delta = 10$ and $\lambda = 0.005$ for WN11, $\delta = 5$ and $\lambda = 0.01$ for FB13, and $\delta = 5$ and $\lambda = 0.005$ for NELL186. For the relation prediction task, 
we selected $\delta = 10$ and $\lambda = 0.005$ for WN11, $\delta = 10$ and $\lambda = 0.05$ for FB13, and $\delta = 1$ and $\lambda = 0.05$ for NELL186.

\begin{table*}[!ht]
\centering
\begin{tabular}{l|l|l|ll|lll|lll}
\hline
\multirow{2}{*}{\textbf{Data}} & \multicolumn{2}{|c|}{{\multirow{2}{*}{\textbf{Method}}}} & \multicolumn{2}{|c|}{\textbf{Triple class.}} & \multicolumn{3}{|c|}{\textbf{Entity prediction}} & \multicolumn{3}{|c}{\textbf{Relation prediction}}\\
\cline{4-11}
 & \multicolumn{2}{|c|}{} & Mic. & Mac. & MR &  MRR &  H@10 & MR &  MRR    &  H@10 \\
\hline
\multirow{4}{*}{WN11} & \multirow{2}{*}{R} & TransE   & 85.21 & 82.53 &  4324 & \textbf{0.102} & \textbf{19.21} & 2.37 & 0.679  & \textbf{99.93}  \\
& &   TransE-NMM  &  \textbf{86.82} & \textbf{84.37} & \textbf{3687}   & 0.094 & 17.98 & \textbf{2.14} & \textbf{0.687} & 99.92 \\
\cline{3-11}
\cline{2-11}
& \multirow{2}{*}{F} & TransE & \multicolumn{2}{c|}{\multirow{2}{*}{}} &   4304 & \textbf{0.122} & \textbf{21.86} & 2.37 & 0.679  & \textbf{99.93} \\
& &  TransE-NMM  &  & &   \textbf{3668}  & 0.109 & 20.12 & \textbf{2.14} & \textbf{0.687} & 99.92\\
\cline{3-3}\cline{6-11}

\hline
\hline

\multirow{4}{*}{FB13} & \multirow{2}{*}{R} & TransE   &  87.57 & 86.66 &  9037   & 0.204  & 35.39  & 1.01 & 0.996 & 99.99    \\
& &   TransE-NMM  & \textbf{88.58} & \textbf{87.99} &  \textbf{8289}   & \textbf{0.258}  &  \textbf{35.53} & 1.01 & 0.996 & \textbf{100.0} \\
\cline{3-11}
\cline{2-11}
& \multirow{2}{*}{F} & TransE & \multicolumn{2}{c|}{\multirow{2}{*}{}} & 5600   & 0.213  &  36.28 & 1.01 & 0.996 & 99.99    \\
& &  TransE-NMM  & & & \textbf{5018} & \textbf{0.267} &  \textbf{36.36} & 1.01 & 0.996 & \textbf{100.0} \\
\cline{3-3}\cline{6-11}
\hline
\hline

\multirow{4}{*}{NELL186} & \multirow{2}{*}{R} & TransE   &  92.13 & 88.96 &   309 & 0.192 & 36.55 &  8.43 & 0.580 & 77.18 \\
& &   TransE-NMM  &  \textbf{94.57} & \textbf{90.95} &  \textbf{238} & \textbf{0.221} & \textbf{37.55} &  \textbf{6.15} & \textbf{0.677} & \textbf{82.16} \\
\cline{3-11}
\cline{2-11}
& \multirow{2}{*}{F} & TransE & \multicolumn{2}{c|}{\multirow{2}{*}{}} &  279 & 0.268 &  47.13 &  8.32 & 0.602 & 77.26   \\
& &  TransE-NMM  &  & & \textbf{214} & \textbf{0.292} & \textbf{47.82} &  \textbf{6.08} & \textbf{0.690} & \textbf{82.20} \\
\cline{3-3}\cline{6-11}
\hline
\end{tabular}
\caption{Experimental results of TransE (i.e. TransE-NMM with $\tau = 0$) and TransE-NMM with $\tau = 10$. Micro-averaged (labeled as \textbf{Mic.}) and Macro-averaged (labeled as \textbf{Mac.}) accuracy results are for the triple classification task.  MR, MRR and H@10 abbreviate the mean rank, the mean reciprocal rank  and Hits@10 (in \%), respectively. ``R'' and ``F'' denote the ``Raw'' and ``Filtered'' settings used in the entity prediction and relation prediction tasks, respectively.} 
\label{tab:results}
\end{table*}

\section{Results}

\subsection{Quantitative results}
Table \ref{tab:results} presents the results of TransE and TransE-NMM  on triple classification, entity prediction and relation prediction tasks on all experimental datasets. The results show that TransE-NMM generally performs better than TransE in all three evaluation tasks. 

\begin{table}[!ht]
\centering
\resizebox{7.75cm}{!}{
\begin{tabular}{l|ll}
\hline
\bf Method &\bf W11 & \bf F13  \\
\hline
TransR \cite{AAAI159571} & 85.9 & 82.5 \\
CTransR \cite{AAAI159571} & 85.7 & - \\
TEKE\_H  \cite{DBLP:conf/ijcai/WangL16}  & 84.8 & 84.2 \\
TransD \cite{ji-EtAl:2015:ACL-IJCNLP} & {86.4} & \textbf{89.1} \\
TranSparse-S \cite{JiLH016} & {86.4} & 88.2 \\
TranSparse-US \cite{JiLH016} & \underline{86.8} & 87.5 \\
\hline
NTN \cite{NIPS2013_5028} & 70.6 & 87.2\\
TransH \cite{AAAI148531} & 78.8 & 83.3 \\
SLogAn \cite{LiangF15} & 75.3 & 85.3 \\
KG2E \cite{He:2015} & 85.4 & 85.3   \\
Bilinear-\textsc{comp} \cite{guu-miller-liang:2015:EMNLP} & 77.6 & 86.1\\
TransE-\textsc{comp} \cite{guu-miller-liang:2015:EMNLP} & 80.3 & 87.6 \\
TransR-FT \cite{FengHWZHZ16} & 86.6 & 82.9 \\
TransG \cite{xiao-huang-zhu:2016:P16-1} & \textbf{87.4} & 87.3 \\
lppTransD \cite{yoon-EtAl:2016:N16-1}  & 86.2 & \underline{88.6} \\
\hline
TransE & {85.2} & {87.6}  \\
\hline
TransE-NMM & \underline{86.8} & \underline{88.6}\\
\hline
\end{tabular}
}
\caption{Micro-averaged accuracy results (in \%) for triple classification on WN11 (labeled as \textbf{W11}) and FB13 (labeled as \textbf{F13}) test sets. Scores in \textbf{bold} and \underline{underline} are the best
and second best scores, respectively.}
\vspace{-5pt}
\label{tab:compared1}
\end{table}

\begin{table}[!ht]
\centering
\resizebox{7.75cm}{!}{
\def\arraystretch{1.125}
\begin{tabular}{l|ll|ll}
\hline
\multirow{2}{*}{\bf Method} & \multicolumn{2}{|c|}{\bf Triple class.} & \multicolumn{2}{|c}{\bf Entity pred.} \\
\cline{2-5}
 &  Mic. &  Mac. & MR & H@10 \\
\hline
TransE-LLE  & 90.08 & 84.50 & 535 & 20.02 \\
SME-LLE & 93.64 & 89.39 & \underline{253}   &  37.14 \\
SE-LLE & \underline{93.95} & 88.54 & 447   &  31.55 \\
\hline
TransE-SkipG & 85.33  & 80.06 & 385  & 30.52 \\
SME-SkipG &  92.86   &  \underline{89.65} &  293   &  \textbf{39.70} \\
SE-SkipG  &  93.07   &  87.98 &  412   &  31.12  \\
\hline
TransE  & {92.13} & {88.96} & 309 & 36.55 \\
\hline
  TransE-NMM  & \textbf{94.57} & {\textbf{90.95}} & \textbf{238} & \underline{37.55}\\
\hline 
\end{tabular}
}
\caption{Results on   the NELL186 test set. Results for the entity prediction task are in the ``Raw'' setting. ``-SkipG'' abbreviates ``-Skip-gram''.}
\vspace{-5pt}
\label{tab:compared2}
\end{table}

Specifically, TransE-NMM obtains higher triple classification results than  TransE in all three experimental datasets, for example, with 2.44\% absolute improvement  in the micro-averaged accuracy on the NELL186 dataset (i.e. 31\% reduction in error). In terms of entity prediction, TransE-NMM obtains better mean rank, MRR and Hits@10 scores than TransE on both FB13 and NELL186 datasets. Specifically, on NELL186 TransE-NMM gains a significant improvement of $279 - 214 = 65$ in the filtered mean rank (which is about 23\% relative improvement), while on the FB13 dataset, TransE-NMM improves with $0.267 - 0.213 = 0.054$ in the filtered MRR (which is about 25\% relative improvement). On the WN11 dataset, TransE-NMM only achieves better mean rank for entity prediction. 
 The relation prediction results of TransE-NMM and TransE are relatively similar on both WN11 and FB13 because the number of relation types is small in these two datasets.  On NELL186, however,  TransE-NMM does significantly better than TransE.

In Table \ref{tab:compared1}, we compare the micro-averaged triple classification accuracy  of our  TransE-NMM model with the previously reported results on the WN11 and FB13 datasets. The first 6 rows report the performance of models that use TransE  to initialize the entity and relation vectors.  
The last 11 rows present the accuracy of models with randomly initialized parameters.

Table \ref{tab:compared1} shows that our  TransE-NMM model obtains  the second highest result on both WN11 and FB13.  Note that there are higher results reported for NTN \cite{NIPS2013_5028}, Bilinear-\textsc{comp} and  TransE-\textsc{comp} \cite{guu-miller-liang:2015:EMNLP}  when entity vectors are initialized by averaging the pre-trained word vectors  \cite{Mikolov13b,Pennington14}.
It is not surprising as many entity names in WordNet and FreeBase are lexically meaningful. It is possible for all other embedding models to utilize the pre-trained word vectors as well. However, as pointed out by \newcite{AAAI148531} and \newcite{guu-miller-liang:2015:EMNLP}, averaging the pre-trained word vectors for initializing entity vectors  is an open problem and it is not always useful since entity names in many domain-specific KBs are not lexically meaningful. 

Table \ref{tab:compared2} compares  the accuracy for triple classification, the raw mean rank and raw Hits@10 scores for entity prediction on the NELL186 dataset. The first three rows present the best results reported in \newcite{guo-EtAl:2015:ACL-IJCNLP1}, while the next three rows present the best results  reported in \newcite{luo-EtAl:2015:EMNLP3}. 
TransE-NMM obtains the highest triple classification accuracy, the best raw mean rank and the second highest raw Hits@10 on the entity prediction task in this comparison.

\subsection{Qualitative results}

Table \ref{tab:qualitative} presents some examples to illustrate the useful information modeled by the neighbors. 
We took the relation-specific mixture weights from the learned TransE-NMM model optimized on the entity prediction task, and then extracted three neighbor relations with the largest mixture weights given a relation. 

Table \ref{tab:qualitative} shows that those relations are semantically coherent. For example, if we know the place of birth and/or the place of death of a person  and/or the location where the person is living, it is likely that we can predict the person's nationality. On the other hand, if we know that a person works for an organization and that this person is also the top member of that organization, then it is possible that this person is the CEO of that organization.  
  
\begin{table}
\resizebox{7.75cm}{!}{
\begin{tabular}{l|l}
\hline
\bf Relation & \bf Top 3-neighbor relations \\
\hline
\multirow{2}{*}{ has\_instance} &	type\_of \\
& subordinate\_instance\_of \\
(WN11)& domain\_topic \\
\hline
\multirow{2}{*}{synset\_domain\_topic} &	domain\_region \\
&	member\_holonym \\
(WN11)& member\_meronym \\
\hline
\multirow{2}{*}{nationality} &	place\_of\_birth \\
& 	place\_of\_death \\
(FB13)&  	location \\
\hline
spouse  &	\multirow{2}{*}{children, spouse, parents} \\
(FB13) & \\
\hline
\multirow{2}{*}{CEOof} &	 WorksFor \\
 & TopMemberOfOrganization 	\\
(NELL186) &  PersonLeadsOrganization \\
 \hline
 \multirow{2}{*}{AnimalDevelopDisease} &	 AnimalSuchAsInsect \\
 & AnimalThatFeedOnInsect 	\\
(NELL186) &  AnimalDevelopDisease \\
\hline
\end{tabular}
}
\caption{Qualitative examples.}
\label{tab:qualitative}
\end{table}

\subsection{Discussion}

Despite of the lower triple classification scores of TransE reported in \newcite{AAAI148531}, Table~\ref{tab:compared1} shows that TransE in fact obtains a very competitive accuracy. 
 Particularly, compared to the relation path model TransE-\textsc{comp} \cite{guu-miller-liang:2015:EMNLP}, when model parameters were randomly initialized, TransE obtains $85.2 - 80.3 = 4.9\%$ absolute accuracy improvement on the WN11 dataset while achieving similar score on the FB13 dataset. Our high results of the TransE model are probably due to a careful grid search and using the ``Bernoulli'' trick.   
  Note that  \newcite{AAAI159571},  \newcite{ji-EtAl:2015:ACL-IJCNLP} and 
 \newcite{JiLH016} did not report the TransE results used for initializing TransR, TransD and TranSparse, respectively. They directly copied the TransE results previously reported in   \newcite{AAAI148531}. So it is difficult to determine exactly how much TransR, TransD and TranSparse gain over TransE. 
 These models might obtain better results than previously reported when the TransE used for initalization performs as well as reported in this paper.
 Furthermore, \newcite{garciaduran-bordes-usunier:2015:EMNLP}, \newcite{lin-EtAl:2015:EMNLP1}, \newcite{Garcia-DuranBUG15} and \newcite{Nickel:2016:HEK:3016100.3016172} also  showed that for entity prediction TransE obtains very competitive results which are much higher than the TransE results originally published  in \newcite{NIPS2013_5071}.\footnote{They did not report the results on WN11 and FB13 datasets, which are used in this paper, though.}

 \begin{figure}[t]
\centering
\includegraphics[width=7.5cm]{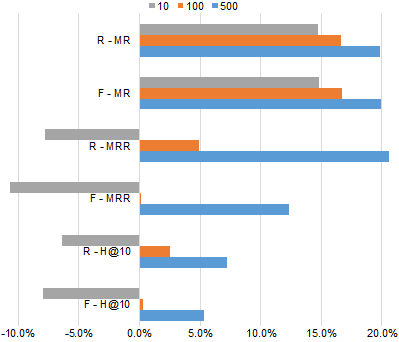}
\caption{Relative improvement of TransE-NMM against TransE for entity prediction task in WN11 when the filtering threshold 
$\tau = \{10, 100, 500\}$ (with other hyper-parameters being the same as selected in Section \ref{ssec:hyper}). 
Prefixes ``R-'' and ``F-'' denote the ``Raw'' and ``Filtered'' settings, respectively. Suffixes ``-MR'', ``-MRR'' and ``-H@10'' abbreviate the mean rank, the mean reciprocal rank  and Hits@10, respectively. }
\label{fig:WN11_results}
\end{figure}

As presented in Table \ref{tab:results}, for entity prediction using  WN11, TransE-NMM with the filtering threshold $\tau=10$ only obtains better mean rank than TransE (about 15\% relative improvement) but lower Hits@10 and mean reciprocal rank. The reason might be that in semantic lexical KBs such as WordNet where relationships between words or word groups are manually constructed, whole neighborhood information might be useful. So when using a small filtering threshold, the model ignores a lot of  potential information that could help predicting relationships. Figure \ref{fig:WN11_results} presents relative improvements in entity prediction of TransE-NMM over TransE on WN11 when varying the filtering threshold $\tau$. Figure \ref{fig:WN11_results}  shows that TransE-NMM gains better scores with higher $\tau$ value. Specifically, when  $\tau = 500$ TransE-NMM does significantly better than TransE in all entity prediction metrics.

\section{Conclusion and future work}

We introduced a neighborhood mixture model for knowledge base completion by constructing neighbor-based vector representations for entities. We demonstrated its effect by extending TransE \cite{NIPS2013_5071} with our neighborhood mixture model. On three different datasets, experimental results show that our model significantly improves TransE and obtains better results than the other state-of-the-art embedding models on triple classification, entity prediction and relation prediction tasks. In future work, we plan to apply the neighborhood mixture model to other embedding models, especially to relation path models such as TransE-\textsc{comp}, to combine the useful information from both relation paths and entity neighborhoods.   

\section*{Acknowledgments}  
This research was supported by a Google award through the Natural 
Language Understanding Focused Program, and under the Australian 
Research Council's {\em Discovery Projects} funding scheme (project 
number DP160102156). This research was also supported by NICTA, funded by the Australian Government through the Department of Communications and the Australian Research Council through the ICT Centre of Excellence Program. 
The first author was supported by an International Postgraduate Research Scholarship and a NICTA NRPA Top-Up Scholarship.

\bibliographystyle{acl2016}
\bibliography{REFs}

\end{document}